# TACO – Twitter Arguments from COnversations

**Marc Feger[1], Stefan Dietze[2]**
[1]HeiCAD - Heine Center for Artificial Intelligence and Data Science,
[2]GESIS - Leibniz Institute for the Social Sciences
[1]Düsseldorf Germany, [2]Cologne Germany
marc.feger@hhu.de, stefan.dietze@gesis.org

**Abstract**

Twitter has emerged as a global hub for engaging in online conversations and as a research corpus for various disciplines that have recognized the significance of its user-generated content. Argument mining is an important analytical task for processing and understanding online discourse. Specifically, it aims to identify the structural elements of arguments, denoted as information and inference. These elements, however, are not static and may require context within the conversation they are in, yet there is a lack of data and annotation frameworks addressing this dynamic aspect on Twitter. We contribute *TACO*, the first dataset of *T*witter *A*rguments utilizing 1,814 tweets covering 200 entire *CO*nversations spanning six heterogeneous topics annotated with an agreement of 0.718 Krippendorff's $\alpha$ among six experts. Second, we provide our annotation framework, incorporating definitions from the Cambridge Dictionary, to define and identify argument components on Twitter. Our transformer-based classifier achieves an 85.06% macro F1 baseline score in detecting arguments. Moreover, our data reveals that Twitter users tend to engage in discussions involving informed inferences and information. TACO serves multiple purposes, such as training tweet classifiers to manage tweets based on inference and information elements, while also providing valuable insights into the conversational reply patterns of tweets.

## 1. Introduction

Social media has created an open network of voices, connecting people globally and allowing them to exchange ideas and engage in discussions on any topic of interest. Despite these benefits, maintaining healthy and substantial online deliberation and promoting transparent information exchange remain key challenges in this area (Chadwick and Howard, 2008; Ruiz et al., 2011). Twitter, now X, serves as a global hub for opinions, news, and information, recognized for its research value and user-generated content prior to its rebranding (Kwak et al., 2010; Boyd et al., 2010).

In this context, argument mining has emerged as a valuable technique to identify the structure of inference and reasoning presented as arguments in natural language and is closely related to information extraction, fact checking, citation and opinion mining (Lawrence and Reed, 2019). This involves the automatic identification and extraction of arguments expressed in text, thus enabling researchers to analyze and understand the nature and structural elements of online discussions. Over the past years, the field of argument mining has undergone significant development in various domains, such as legal texts (Moens et al., 2007; Wyner et al., 2010), newspapers (Reed et al., 2008; Mochales and Moens, 2011), essays (Stab and Gurevych, 2014; Persing and Ng, 2016; Wachsmuth et al., 2016), Wikipedia articles (Levy et al., 2014, 2017), and sources of conflicting content, such as user comments (Park and Cardie, 2014), dialogues (Swanson et al., 2015), and web discourses (Habernal and Gurevych, 2017). While these works made initial contributions to the field, more recent research has focused on the detection of arguments from heterogeneous sources of arbitrary web text (Daxenberger et al., 2017; Levy et al., 2018; Stab et al., 2018). Despite addressing different aspects of argument mining, all studies involve the detection of inference and information (Palau and Moens, 2009; Daxenberger et al., 2017) as part of online discourse.[1]

| |
|---|
| (1) Men shouldn't be making laws about women's bodies #abortion #Texas |
| (2) 'Bitter truth': EU chief pours cold water on idea of Brits keeping EU citizenship after #Brexit https://t.co/j3DteyWcMg via @TheLocalEurope |
| (3) Opinion: As the draconian (and then some) abortion law takes effect in #Texas, this is not an idle question for millions of Americans. A slippery slope towards more like-minded Republican state legislatures to try to follow suit. #abortion #F24 https://t.co/sMKUdhRF1g |
| (4) @sinnfeinireland Blah blah blah blah blah blah |

Table 1: Example tweets that contain inference (1), information (2), a combination of both (3), or none of either (4).

For argument mining on Twitter, research has expanded from specific topics like football (Llewellyn et al., 2014) and encryption (Addawood and Bashir, 2016) to encompass various subjects, including Brexit and Grexit (Dusmanu et al., 2017). Recent studies have focused on structuring tweets into debates via semantic similarity on the topics Iran, Grexit, Apple Watch and Game of Thrones (Bosc et al., 2016), and with isolated tweet-reply

---

[1]These components, inference and information, are defined along the annotation framework in Section 2.1.

pairs on climate (Schaefer and Stede, 2020), neither capturing entire conversations.

This diversification has led to specialized tasks, such as identifying pro and con arguments in Planned Parenthood tweets (Bhatti et al., 2021) and evaluating scientific support in Covid-19 and climate-related tweets (Hafid et al., 2022).

Despite the progress in argument mining, the scope of related research on Twitter is restricted to a micro-level perspective, solely examining individual tweets and neglecting the interrelated reply tweets that make up the wider context of Twitter discussions. So far, no ground truth data for assessing arguments in entire Twitter conversations exists (Schaefer and Stede, 2021). With our work, we contribute the following to advance the field of argument mining on Twitter:

1. **Annotation Framework.** Our specialized argument mining framework for Twitter conversations evolved from an extensive analysis of the elements defining arguments in relevant literature and iterative deliberations among our experts. It builds on Cambridge Dictionary's[2] definitions to define and identify inference and information within tweets.

2. **Conversation-Based Ground Truth Data.** Our TACO[3] dataset comprises 1,814 tweets, covering 200 entire conversations from six widely-discussed Twitter events. It was annotated by six experts with a high agreement score of 0.718 Krippendorff's α. TACO is available to the public in compliance with Twitter's data policy.

3. **Baseline Classification Model.** Our published transformer-based classifier[4], underlines TACO's significance in argument mining by achieving an 85.06% macro F1 for detecting arguments in tweets and 72.49% macro F1 for identifying combinations of inference and information. This classifier can be employed in both cases to aid in building new datasets and tweet curation.

## 2. Constructing the TACO dataset

Given the brevity of tweets, which were originally limited to 140 respectively 280 characters, structural elements of arguments such as inference or information tend to be scattered across distinct messages (Kwak et al., 2010; Boyd et al., 2010; Addawood and Bashir, 2016; Dusmanu et al., 2017). At the same time, tweets tend to be rather diverse in nature. Some tweets indicate a genuine interest in contributing to ongoing debates, while others may express different motivations, such as a what-i-had-for-lunch-like tweet (Rogers, 2013).

### 2.1. Annotation Framework

With no one-size-fits-all definition of what an argument is (Palau and Moens, 2009; Habernal et al., 2014; Stab et al., 2018), the crucial challenge is how to identify arguments on Twitter. However, one potential strategy for simplifying this task is to differentiate between tweets that contain an inference as a key component of an argument and those that do not (Palau and Moens, 2009; Stab and Gurevych, 2014; Bosc et al., 2016; Daxenberger et al., 2017; Lawrence and Reed, 2019). Our aim here is not to create a new formalism for arguments, but rather to integrate established theories and provide a reusable set of definitions that can be applied to Twitter.

To define this critical component of an argument, we refer to the Cambridge Dictionary, which defines *inference as a guess that you make or an opinion that you form based on the information that you have*. We also utilize their description of *information* as *facts or details about a person, company, product, etc.*.

Taken together, argument mining on Twitter involves determining if a tweet contains an inference (**Argument**) or not (**No-Argument**) by also considering its combination with information, as illustrated in Figure 1.

Tweets categorized as an **Argument** can be:

**Statement**, a tweet where only inference is presented like *something that someone says or writes officially, or an action done to express an opinion* (see tweet 1 in Table 1).

**Reason**, a tweet where the inference is based on information mentioned in the tweet, such as references, and thus reveals the author's motivation *to try to understand and to make judgments based on practical facts* (see tweet 3 in Table 1).

In contrast, tweets that are categorized as **No-Argument** are defined by the absence of inference and can be described as:

**Notification**, a tweet that is limited to only providing information, such as media channels promoting their articles (see tweet 2 in Table 1).

**None**, a tweet that provides neither inference nor information (see tweet 4 in Table 1).

### 2.2. Data Sampling and Annotation

Twitter conversations are shown to have a strong focus on various topics, often driven by hashtags (Hughes and Palen, 2009; Rogers, 2013; Zhou and Chen, 2014). We utilized Twitter's API v2 to collect a corpus of tweets and their reply-relations, enabling the extraction of entire conversations.

---
[2]dictionary.cambridge.org
[3]github.com/TomatenMarc/TACO
[4]huggingface.co/TomatenMarc/TACO

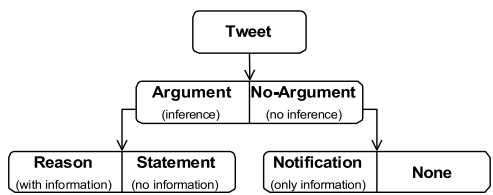

Figure 1: Hierarchy of inference and information.

While earlier studies of argument mining on Twitter mostly focused on one or two topics (Llewellyn et al., 2014; Addawood and Bashir, 2016; Dusmanu et al., 2017; Schaefer and Stede, 2020; Bhatti et al., 2021; Hafid et al., 2022), we aimed to create a more comprehensive corpus by collecting tweets from six controversial topics: #Abortion, #Brexit, #GOT, #TwitterTakeover, #SquidGame, and #LOTRROP. Over a period of seven months, we collected ~600k tweets around key-dates[5] related to these hashtags, resulting in a significant increase in tweet volume. The allocation of the ~600k tweets and the profiles of the hashtag topics are as follows:[5]

**#Abortion** (5%) pertaining to the 'Row v. Wade' lawsuit, which challenges the Texas abortion ban (S.B.8) after six weeks, discussed and obtained between August 15 and October 16, 2021.

**#Brexit** (70.9%) relates to the global discussion of Great Britain's departure from the European Union on February 1, 2020. This historic event was queried from January 1 to March 1, 2020.

**#GOT** (10.2%) was gathered from April 1 to May 1, 2019, for expressing criticism about the final season of 'Game of Thrones'.

**#TwitterTakeover** (3.1%) was queried from April 1 to May 1, 2022, in association with Elon Musk's position as Twitter's largest shareholder, raising concerns about freedom of speech and initiating its transition to X.

**#SquidGame** (8.5%) explores economic inequality's impact on moral choices while also advertising the corresponding Netflix series, collected from September 10 to October 10, 2021.

**#LOTRROP** (2.3%) discussed the release of the trailer for Amazon's first season of 'The Lord of the Rings: The Rings of Power' and the related debate on representation and diversity in media, which we tracked from February 1 to March 1, 2022.

Text annotation is a multifaceted task, encompassing reading, comparing, memorization, and developing consensus about the data (Guetterman et al., 2018; Geiger et al., 2020). Under the author's guidance, two annotation rounds were performed with distinct expert groups to classify tweets based on our proposed annotation framework for Twitter conversations.[5]

---

[5]For the meta-data or examples, see: README.md

In step 1, three experts and the paper's first author refined the framework's guidelines and assessed their generalizability across topics. We sampled 300 conversation-starting tweets for #Abortion and #Brexit, given their significance in argument mining (Wachsmuth et al., 2017a; Dusmanu et al., 2017; Stab et al., 2018; Levy et al., 2018; Bhatti et al., 2021). Only when complete agreement on a tweet's classification was reached did it proceed as a candidate for step 2 annotation.

In step 2, the first author and two additional experts annotated 200 full conversations, tracing the sequences of reply-relations from the starting tweets to the final replies in a conversation, thereby considering the conversational context.[5] This included 50 conversations for both #Abortion and #Brexit, and 25 conversations for each of the remaining four topics. In total, 236 #Abortion, 285 #Brexit, 192 #GOT, 166 #TwitterTakeover, 226 #SquidGame, and 209 #LOTRROP tweets were considered, averaging 219 tweets per topic.

In these steps, 1,814 annotated tweets were generated, including 1,314 conversation-based tweets of all topics and 500 distinct conversation-starting tweets for #Abortion and #Brexit, with a strong 0.718 Krippendorff's $\alpha$ agreement, given the author's involvement in both phases.

### 2.3. TACO Dataset

The final TACO ground truth (see Table 2) involved a strict majority vote approach, discarding tweets with less than 50% agreement among assigned votes for a specific class, resulting in 1,734 tweets, accounting for 95.6% of the annotated tweets.

| Category | Argument | | No-Argument | |
|---|---|---|---|---|
| | 865 (49.88%) | | 869 (50.12%) | |
| Class | Reason | Statement | Notification | None |
| | 581 (33.50%) | 284 (16.38%) | 500 (28.84%) | 369 (21.28%) |

Table 2: Class distribution in the TACO dataset.

## 3. Argument Mining on Twitter

We trained a soft-max classifier on top of different transformer models to utilize TACO in two tasks: (1) detecting inference in tweets (**Argument** vs. **No-Argument**), and (2) classifying tweets based on combinations of inference and information.

To obtain a first baseline for the usability of TACO, we fine-tuned the classifiers within an ordinary sequence classification approach integrating TACO's tweets, whose labels, created using our conversation-based annotation framework, encode implicit conversational context without detailing the conversation structures.[6]

---

[6]This approach is pragmatic, avoiding the complexity of modeling conversation hierarchies during fine-tuning.

For task (2), we fine-tuned the classifiers using 10-fold cross-validation, whereby the experimental results showed that BERTweet (Nguyen et al., 2020) provides superior classification performance. Further it is worth noting that the results for task (1) are aggregations of task (2), with a focus on presence or absence of inference.

The benefits of the BERTweet classifier for TACO extend beyond theory and are supported by cross-validation demonstrating strong performance for argument mining across the conversation-based tweets. Our results demonstrate this effectiveness, with the baseline model achieving an 85.06% macro F1 for inference detection of task (1) and 72.49% for classification of task (2), as seen in Table 3.

In terms of TACO's data, the classification model had access to the following text features indicating tweet classes. The length of tweets differs among classes, with Reason being the longest on average (213), None the shortest (63), and Notification (156) and Statement (122) falling in the middle character range. URL usage varies, with Reason (34.6%) and Notification (71.6%) having the highest, while None and Statement use them less than 8.11%. The usage of discourse marker[7] aligns with the argumentativeness of tweets: Reason (32.9%) is highest, followed by Statement (19%), Notification (11.4%), and None (8.7%).

Although there are misclassifications in task (2), ~43% of them are counted as true positives in task (1). The misclassification of Reason as Statement (or vice versa) still falls into the category of **Argument**, the same being true for the mutual misclassification of Notification and None, which still contributes as **No-Argument**.

Besides inference detection, the model also faces the challenge of correctly classifying tweets by identifying the informative parts, which is distinct from detecting inference and may not be entirely satisfactory due to the varied forms in which information can be presented, like URLs or fragmented text passages. Different tweets may employ language for specific intents involving rhetorical devices or visual elements, adding complexity to identifying information's formal attributes.

Error analysis revealed that Statements, although typically lacking information, can include URLs when misclassified as Reason or Notification, as URLs might seem like apparent markers of information, resulting in an increased average length of 172. In fact, 22.09% of these misclassified Statements contained URLs, predominantly from internal sources like memes, GIFs, and videos. This circumstance might be influenced by Twitter's recommendation system, which rewards

---
[7] dictionary.cambridge.org/us/grammar/british-grammar/discourse-markers

| Task | Instance | Precision (%) | Recall (%) | F1 (%) | macro F1 (%) |
|---|---|---|---|---|---|
| Category | Argument | 83.59 | **87.17** | **85.34** | 85.06 |
| | No-Argument | **86.66** | 82.97 | 84.77 | |
| Class | Reason | 73.69 | 75.22 | 74.45 | 72.49 |
| | Statement | 54.37 | 59.15 | 56.66 | |
| | Notification | 79.02 | **77.60** | 78.30 | |
| | None | **83.87** | 77.51 | **80.56** | |

Table 3: Baseline argument mining performance.

tweets that attract a large audience and have entertaining contents attached.[8]

Further investigation found that 24.42% of these incorrect assignments in the Statement class contained discourse markers, which complicated classification because these markers often organized relations between information and inference, leading to multiple stacked inferences that amplified the message's tone rather than being perceived as information in the first instance.

To provide context, the model achieved a F1 score of 80.56% in detecting tweets lacking inference and information (None), which often led to conversation halts in about one-third (33.15%) of cases where a tweet received no further replies. Reason was the second most common in ending conversations at a rate of 29.73%, potentially indicating knockout arguments of further interest. Additionally, a Statement received no replies in 19.04% of cases, while Notification had no replies in 18.08%.

### 3.1. Conversational Reply Patterns

In our final analysis of the conversation-based ground truth data, we explored state transitions between connected tweet-reply pairs $(t, r)$ to reveal TACO's value in understanding reply patterns and providing insights into conversation progression, as shown in Table 4.

| $P(r|t)$ | Reason | Statement | Notification | None |
|---|---|---|---|---|
| **Reason** | **0.51** | 0.12 | 0.31 | 0.06 |
| **Statement** | **0.38** | 0.21 | 0.33 | 0.08 |
| **Notification** | 0.26 | 0.08 | **0.57** | 0.09 |
| **None** | 0.26 | 0.08 | **0.44** | 0.22 |

Table 4: Transition probability of a tweet with class $t$ (row) having a reply with class $r$ (column).

Our findings reveal users often reply with informed inferences (Reason) or additional information (Notification), with less common conversations solely based on inference (Statement) or lacking both elements (None). Additionally, **Argument** relies on informed inference, while **No-Argument** depends on information usage in replies, reflecting a preference for informed debates.

---
[8] github.com/twitter/the-algorithm

## 4. Conclusion

This paper presents the first ground truth dataset, TACO, for conversation-based argument mining on Twitter, efficiently annotated by six experts using our purpose-built annotation framework that incorporates definitions of argument constituting elements from the Cambridge Dictionary. Unlike previous datasets that often rely on isolated tweets without the contextual framework of conversations, TACO offers fully annotated and coherent Twitter conversations across six topics used for training a transformer-based classification model, providing valuable resources for future research in this domain. Furthermore, the provided classifier effectively differentiates tweets that make arguments from those that do not, based on the presence of inference. Additionally, our multi-class approach sufficiently identifies tweet classes, especially those that lack information and inference. Our findings suggest the need for further research to enhance the semantic features of our proposed tweet classes, possibly by fine-tuning BERTweet for more generalized representations according to our framework.

## 5. Ethics Statement

In the context of this study, which uses Twitter data, we have adhered to ethical practices and privacy principles and ensured data protection by limiting the publication of TACO to tweet IDs in accordance with Twitter's terms of service. Our annotation process, which involved volunteer experts, has been carefully designed to limit data access to what was strictly necessary and to ensure ethical standards, fair compensation and data integrity. Access to the original dataset is restricted to non-harmful research, subject to appropriate data protection agreements with the authors. It should also be noted that the TACO dataset covers sensitive topics that may contain language and images that some may find offensive.

## 6. Acknowledgments

## 7. Bibliographical References